\documentclass[9pt,conference]{IEEEtran}
\usepackage{amssymb,amsthm,amsmath,array}
\usepackage{graphicx}
\usepackage[caption=false,font=footnotesize]{subfig}
\usepackage{xspace}
\usepackage[sort&compress, numbers]{natbib}
\usepackage{stmaryrd}
\usepackage{xcolor}
\usepackage{mathtools}
\usepackage{float}
\usepackage{textcomp}
\usepackage{caption} 
\usepackage{multirow}
\usepackage{booktabs}
\usepackage{hyperref}
\begin{document}
\title{A Central Asian Food Dataset for Personalized Dietary Interventions, Extended Abstract}
\author{\IEEEauthorblockN{
        Aknur Karabay\IEEEauthorrefmark{1}, 
        Arman Bolatov\IEEEauthorrefmark{1},
        Huseyin Atakan Varol\IEEEauthorrefmark{1}, and
        Mei-Yen Chan\IEEEauthorrefmark{2}
    }
    \IEEEauthorblockA{
        \IEEEauthorrefmark{1} Institute of Smart Systems and Artificial Intelligence, Nazarbayev University.\\
        \IEEEauthorrefmark{2} School of Medicine, Nazarbayev University.\\}
}
\maketitle
\begin{abstract}
    Nowadays, it is common for people to take photographs of every beverage, snack, or meal they eat and then post these photographs on social media platforms. Leveraging these social trends, real-time food recognition and reliable classification of these captured food images can potentially help replace some of the tedious recording and coding of food diaries to enable personalized dietary interventions. Although Central Asian cuisine is culturally and historically distinct, there has been little published data on the food and dietary habits of people in this region. To fill this gap, we aim to create a reliable dataset of regional foods that is easily accessible to both public consumers and researchers. To the best of our knowledge, this is the first work on creating a Central Asian Food Dataset (CAFD). The final dataset contains 42 food categories and over 16,000 images of national dishes unique to this region. We achieved a classification accuracy of 88.70\% (42 classes) on the CAFD using the ResNet152 neural network model. The food recognition models trained on the CAFD demonstrate computer vision's effectiveness and high accuracy for dietary assessment.
\end{abstract}

\section{Introduction}

This manuscript is an extended abstract of our previously published article, A Central Asian Food Dataset for Personalized Dietary Interventions, which appeared in the Nutrients, MDPI in 2023 \cite{nu15071728}.

Food computation from visual data has gained prominence due to computer vision advancements and increased smartphone, and social media usage \cite{Allegra2020}. These platforms provide access to food-related information, which can be utilized for various tasks, including medical, gastronomic, and agronomic research. Deep learning-based food image recognition systems have been developed for applications in dietary assessment, smart restaurants, food safety inspection, and agriculture. Automatic food image recognition can improve the accuracy of nutritional records and assist visually impaired individuals \cite{Allegra2020}.

\begin{table}[H]
\caption{Summary of food classification datasets.}
\centering
\resizebox{\columnwidth}{!}{
    \begin{tabular}{c|c|c|c|c|c}
        \textbf{Dataset} & \textbf{Year} & \textbf{\# class} & \textbf{\# images} & \textbf{Cuisine} & \textbf{Public}\\
        \toprule
        Food-101~\cite{Bossard2014} & 2014 & 101 & 101,000 & European & yes\\
        VireoFood-172~\cite{Chen2016Vireo} & 2016 & 172 & 110,241 & Chinese/Asian & yes\\
        TurkishFoods-15~\cite{Gungor2017} & 2017 & 15 & 7,500 & Turkish & yes\\
        FoodAI~\cite{Sahoo2019} & 2019 & 756 & 400,000 & International & no\\
        VireoFood-251~\cite{Chen2021} & 2020 & 251 & 169,673 & Chinese/Asian & yes\\
        ISIA Food-500~\cite{Min2020ISIAFA} & 2020 & 500 & 399,726 & Chinese/Intern. & yes\\
        Food2K~\cite{Min2021Food2k} & 2021 & 2,000 & 1,036,564 & Chinese/Intern. & no\\
        Food1K~\cite{Min2021Food2k} & 2021 & 1,000 & 400,000 & Chinese/Intern. & yes\\
        \textbf{\textbf{CAFD}}~\cite{nu15071728} &  \textbf{2022} &  \textbf{42 }&  \textbf{16,499} &  \textbf{Central Asian} & \textbf{yes}\\
\end{tabular}}
\label{tab:datasets_summary}
\end{table}

Existing food classification datasets mostly include Western, European, Chinese, and other Asian cuisines \cite{Herzig2020, Sahoo2019}. Examples of such datasets are presented in Table \ref{tab:datasets_summary}. However, FoodAI is not open source, and Food2K is not publicly available. Food1K, a food recognition challenge dataset, has been released with 400,000 images and 1,000 food classes.

Most food datasets predominantly contain Western and Asian dishes, lacking specific national dishes like those in Central Asia. To address this, we aim to develop a unique food recognition system for our region, considering local preferences, specialties, and cuisines. In this paper, we describe the development of our dataset, food recognition models, and their performance and conclude with a summary of our findings.

\section{Central Asian Food Dataset}

This paper introduces the Central Asian Food Dataset (CAFD), consisting of 16,499 images across 42 classes representing popular Central Asian cuisine. We ensured the dataset's high quality through extensive data cleaning, iterative annotation, and multiple inspections. The CAFD can also be a food image representation learning benchmark.

We followed a five-step process to create a diverse and high-quality dataset. First, we listed popular Central Asian food items. Second, we scraped images from search engines and social media using a Python script with Selenium. We extracted images from recipe videos using Roboflow~\cite{Roboflow} to increase underrepresented classes. Third, we removed duplicates using the HashImage Python library. Fourth, two annotators created bounding boxes for each food item using Roboflow software. Fifth, we cropped the food items based on bounding box coordinates and stored them in separate directories by class. The final dataset has an imbalanced number of images per class, ranging from 99 to 922.

\begin{figure}[H]
    \begin{center}
    \includegraphics[width=1.0\columnwidth]{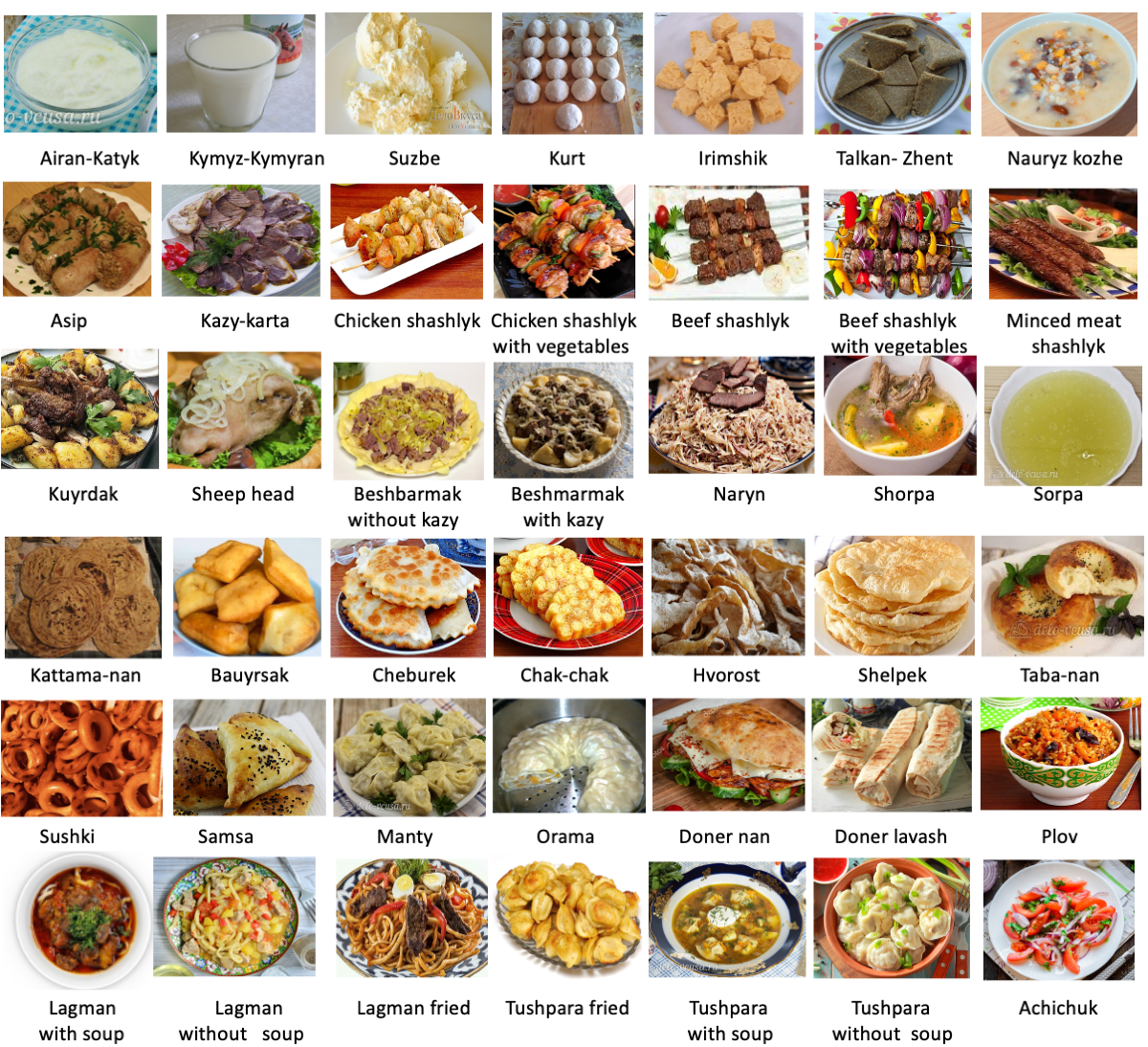}
    \end{center}
    \caption{Sample images for Central Asian Food Dataset classes.}\label{fig:1}
    \label{fig:samples}
\end{figure}

\section{Food Recognition Models}

Image classification is a computer vision task that extracts a single descriptor from an image. State-of-the-art models are based on CNNs and have improved due to large datasets. Transfer learning is often used when sufficient training data is unavailable, as it leverages knowledge from pre-trained models to solve similar problems in different domains~\cite{Torrey2010}. In this work, we applied transfer learning to food classification using model weights pre-trained on ImageNet, a large dataset with over 14 million images~\cite{Deng2009}. 

We selected 10 models of different architectures and complexity to evaluate their performance on the CAFD. These models include VGG-16, Squeezenet1, and five models with ResNet architecture~\cite{Simonyan2014, Iandola2016, He2016, Xie2016, Zagoruyko2016}. DenseNet-121 and EfficientNet-b4 have similar architectures to ResNets but introduce different scaling methods~\cite{Huang2016, Tan2019}. 

Then we trained the models on the Food1K dataset and tested the combination of CAFD and Food1K. We carefully split the datasets into training, validation, and test sets to avoid bias and data leakage. Table~\ref{tab:dataset_sizes} shows the number of images in each set for three different datasets.

Also, we performed transfer learning on Pytorch using pre-trained models on ImageNet. Models were trained for 40 epochs with a learning rate of 0.001, batch size of 64, and a categorical cross-entropy loss. The input size of images varied depending on the model. We used Top-5 accuracy and Top-1 accuracy as evaluation metrics and precision, recall, and $F_1$-score metrics to identify and analyze the best and worst-classified food classes.

\begin{table}[H]
    \caption{Image distribution across the training (train), validation (valid), and test sets.}
    \centering
    \begin{tabular}{c|c|c|c}
        \textbf{Dataset} & \textbf{Train size} & \textbf{Validation size} & \textbf{Test size}\\
        \toprule
        CAFD & 11,008 & 2,763 & 2,728\\
        Food1K & 317,277 & 26,495 & 26,495\\
        CAFD+Food1K & 328,285 & 29,258 & 29,223
    \end{tabular}
    \label{tab:dataset_sizes}
\end{table}

\section{Results and Discussion}

Table~\ref{tab:all_sims} summarizes the classification models' results. All models performed better on the CAFD than on Food1K and CAFD+Food1K, indicating the accuracy and cleanness of the CAFD. VGG-16 achieved 86.03\% Top-1 and 98.33\% Top-5 accuracies on the CAFD, while Squeezenet1 had a lower performance. ResNet architectures achieved around 88\% Top-1 and 98\% Top-5 accuracy on the CAFD, with accuracy increasing as network depth increases. Wide ResNet-50 improved accuracy compared to ResNet50, and EfficientNet-b4 achieved the best results on Food1K and CAFD+Food1K.

\begin{table}[H]
    \caption{Top-1 and Top-5 accuracies for different food classification models and datasets.}
    \centering
    \resizebox{\columnwidth}{!}{%
    \begin{tabular}{c|cc|cc|cc}
        \multirow{2}{*}{\textbf{Base Model}}  &  \multicolumn{2}{c}{\textbf{CAFD}} & \multicolumn{2}{c}{\textbf{Food1k}} & \multicolumn{2}{c}{\textbf{CAFD+Food1K}}\\
        & Top-1 & Top-5 & Top-1 & Top-5 & Top-1 & Top-5\\ \toprule
         VGG-16 & 86.03 & 98.33 & 80.67 & 95.24 & 80.87 & 96.19\\
        Squeezenet1\_0 & 79.58 & 97.29 & 71.33 & 91.23 & 69.16 & 90.15 \\ 
        ResNet50 & 88.03 & 98.44 & 82.44 & 97.01 & 83.22 & 97.25 \\ 
        ResNet101 & 88.51 & 98.44 & 84.10 & 97.34 & 84.20 & 97.45 \\ 
        ResNet152 & \textbf{88.70} & \textbf{98.59} & 84.85 & 97.80 & 84.75 & 97.58\\ 
        ResNext50-32 & 87.95 & 98.44 & 81.17 & 96.67 & 84.81 & 97.65\\
        Wide ResNet-50 & 88.21 & 98.59 & 82.20 & 97.28 & 85.27 & 97.81\\ 
        DenseNet-121 & 86.95 & 98.26 & 83.03 & 97.14 & 82.45 & 96.93 \\ 
        EfficientNet-b4 & 81.28 & 97.37 & \textbf{87.47} & \textbf{98.04} & \textbf{87.75} & \textbf{98.01}\\
    \end{tabular}
    }
    \label{tab:all_sims}
\end{table}

\begin{figure}
    \begin{center}
    \includegraphics[width=1.0\columnwidth]{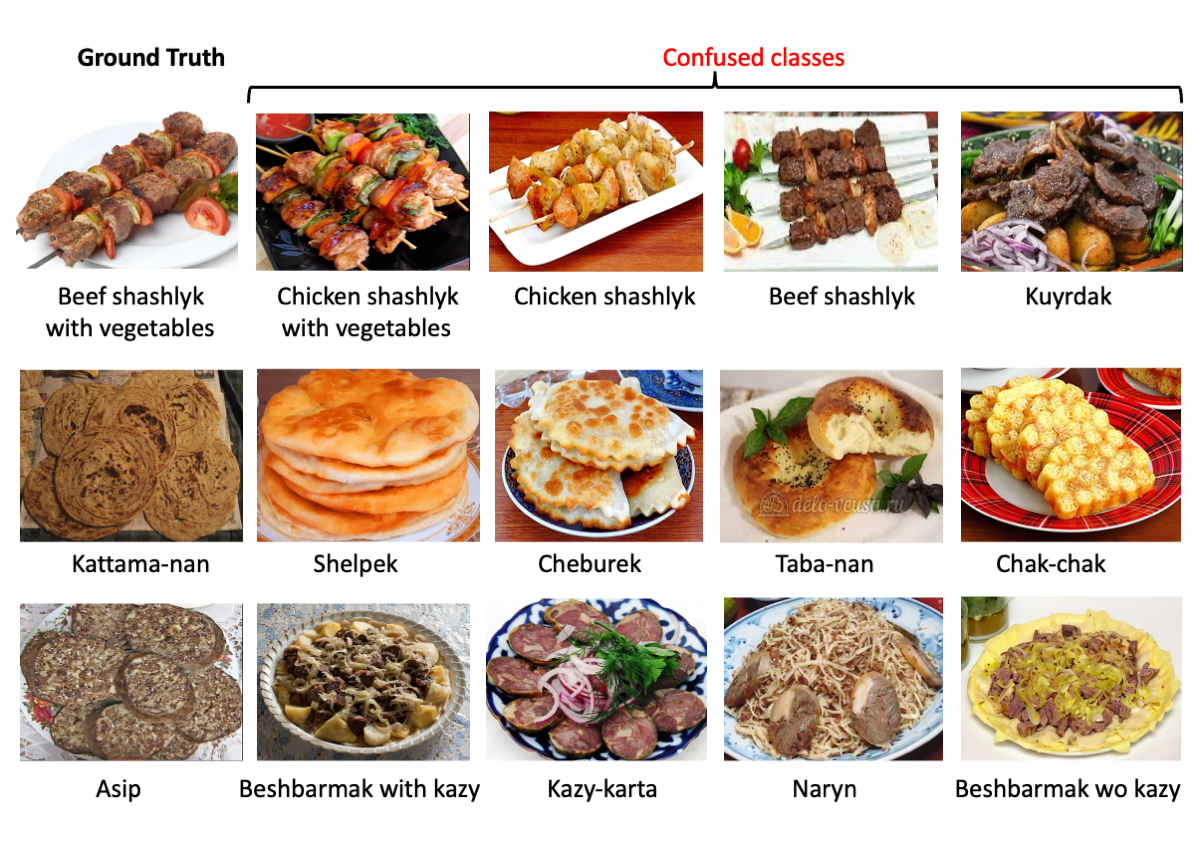}
    \end{center}
    \caption{Examples of the confused classes.}\label{fig:1}
    \label{fig:samples}
\end{figure}

Tables \ref{tab:10_best_worst_CAFD} and \ref{tab:10_best_worst_1k+CAFD} list the 10 best and worst detected CAFD classes by ResNet152 and EfficientNet-b4. Large classes with distinct features performed best, while fine-grained or similar-looking classes, presented in Figure \ref{fig:1}, caused confusion and deteriorated model performance.

\begin{table}[H]
    \caption{Ten CAFD classes best and worst detected by the ResNet152 model.}
    \centering
    \resizebox{\columnwidth}{!}{
        \begin{tabular}{c|c|c|c|c|c|c|c}
            \multicolumn{4}{c}{\textbf{Best detected classes}} & \multicolumn{4}{c}{\textbf{Worst detected classes}} \\
            \textbf{Class} & \textbf{Prec.} & \textbf{Rec.} & \textbf{$F_1$} & \textbf{Class} & \textbf{Prec.} & \textbf{Rec.} & \textbf{$F_1$}\\ \toprule
            Sushki & 0.96 & 1 &  0.98 & Shashlyk chicken w/ v. & 0.71 & 0.67 & 0.69\\
            Achichuk & 0.95 & 1 & 0.98 & Shashlyk beef w/ v. & 0.66 & 0.72 & 0.69\\ 
            Sheep head & 0.94 & 1 & 0.97 & Shashlyk chicken & 0.67 & 0.74 & 0.7\\ 
            Naryn & 0.96 & 0.98 & 0.97 & Shashlyk minced meat & 0.79 & 0.64 & 0.71\\ 
            Plov & 0.93 & 0.99 & 0.96 & Asip & 0.85 & 0.62 & 0.72\\ 
            Tushpara w/ s. & 0.93 & 0.97 & 0.95 & Shashlyk beef & 0.74 & 0.69 & 0.72\\ 
            Soup plain & 0.97 &  0.93 & 0.95 & Lagman without soup & 0.83 & 0.68 & 0.75\\ 
            Samsa & 0.94 & 0.96 & 0.95 & Kazy-karta & 0.83 & 0.74 & 0.78\\ 
            Hvorost & 0.98 & 0.91 & 0.95 & Beshbarmak with kazy & 0.78 & 0.8 & 0.79\\ 
            Manty & 0.92 & 0.95 & 0.94 & Tushpara fried & 0.88 & 0.76 & 0.81\\
    \end{tabular}}
    \label{tab:10_best_worst_CAFD}
\end{table}

\begin{table}[H]
    \caption{Ten CAFD and Food1K classes best and worst detected by the EfficientNet-b4 model.}
    \centering
    \resizebox{\columnwidth}{!}{
        \begin{tabular}{c|c|c|c|c|c|c|c}
            \multicolumn{4}{c}{\textbf{Best detected classes}} & \multicolumn{4}{c}{\textbf{Worst detected classes}} \\
            \textbf{Class} & \textbf{Prec.} & \textbf{Rec.} & \textbf{$F_1$} & \textbf{Class} & \textbf{Prec.} & \textbf{Rec.} & \textbf{$F_1$}\\ \toprule
            Sushki & 0.91 & 1 & 0.96 & Lagman without soup & 0.6 & 0.27 & 0.37\\
            Achichuk & 1 & 0.95 & 0.97 & Asip & 0.88 & 0.38 & 0.53\\
            Sheed head & 0.94 & 0.94 & 0.94 & Talkan-zhent & 0.86 & 0.53 & 0.66\\
            Airan-katyk & 0.83 & 0.93 & 0.88 & Doner lavash & 0.75 & 0.6 & 0.67\\
            Plov & 0.97 & 0.90 & 0.93 & Shashlyk chicken w/ v. & 0.88 & 0.64 & 0.74\\
            Cheburek & 0.92 & 0.90 & 0.91 & Lagman fried & 0.96 & 0.68 & 0.8\\
            Irimshik & 0.93 & 0.88 & 0.91 & Doner nan & 1 & 0.68 & 0.81\\
            Samsa & 0.93 & 0.88 & 0.90 & Shashlyk chicken & 0.61 & 0.69 & 0.65\\
            Naryn & 0.97 & 0.87 & 0.92 & Shahslyk beef & 0.67 & 0.69 & 0.68\\
            Chak-chak & 0.9 & 0.87 & 0.92 & Kazy-karta & 0.8 & 0.7 & 0.74\\
    \end{tabular}}
    \label{tab:10_best_worst_1k+CAFD}
\end{table}

\section{Conclusion}

The Central Asian Food Dataset (CAFD) offers a unique advantage in automating and improving dietary assessment accuracy. It has potential applications in creating or modifying recipes, helping restaurants and food service providers plan menus, optimizing food production, and combating fraudulent food practices. It can be used to improve food quality, develop new recipes and personalized dietary plans, optimize production processes, increase food safety, and integrate with other food recognition systems.

Comprising 16,499 images of 42 food classes, the CAFD demonstrates the effectiveness of computer vision models for food recognition. Our models achieved a Top-5 accuracy of 98.59\% and 98.01\% for the CAFD and CAFD+Food1K, respectively. The dataset, source code, and pre-trained models are available on GitHub
\footnote{\href{https://github.com/IS2AI/Central-Asian-Food-Dataset}{https://github.com/IS2AI/Central-Asian-Food-Dataset}} repository.

Future work includes exploring different neural network architectures, data augmentation methods, and utilizing the CAFD for other dietary-related tasks. We also plan to develop food scene recognition datasets with multiple food items per image and extend the current food categories based on additional food classes.

\section*{Author contributions}

MYC and HAV conceived and designed the study. AK, HAV, and MYC contributed to defining the research scope and objectives. AK and AB collected and prepared the dataset and trained the models. AB created a pipeline for processing images in Roboflow. HAV provided guidelines for the project experiments. AK performed the final check of the dataset and finalized the experimental results. PI of the project: MYC. AK, MYC,and AB wrote the article, and all the authors contributed to the manuscript revision and approved the submitted version.

\bibliographystyle{IEEEtran}
\bibliography{references}

\end{document}